\pdfoutput=1
\documentclass[11pt]{article}

\usepackage[preprint]{acl}
\usepackage{times}
\usepackage{latexsym}
\usepackage[T1]{fontenc}
\usepackage[utf8]{inputenc}
\usepackage{microtype}
\usepackage{inconsolata}
\usepackage{graphicx}
\usepackage{multirow}
\usepackage{enumitem}
\usepackage{amsmath}
\usepackage{booktabs}
\usepackage{amsfonts}
\usepackage{algpseudocode}
\usepackage{algorithm}
\usepackage{setspace}
\usepackage{caption} 
\usepackage{relsize}
\usepackage{ragged2e}
\usepackage{graphicx}
\usepackage{booktabs}
\usepackage{multirow}
\usepackage{algorithm}
\usepackage{algpseudocode}
\usepackage{lipsum}
\usepackage{natbib}
\usepackage{amsmath}
\usepackage{adjustbox}
\usepackage{amssymb}

\usepackage{lipsum}

\usepackage[dvipsnames]{xcolor}
\AtBeginDocument{%
  }
\usepackage{graphicx}
\usepackage{subfig}
\newcommand\scalemath[2]{\scalebox{#1}{\mbox{\ensuremath{\displaystyle #2}}}}
  \PassOptionsToPackage{breaklinks}{hyperref}
  \RequirePackage{hyperref}
  \definecolor{darkblue}{rgb}{0, 0, 0.5}
  \hypersetup{colorlinks=true, citecolor=darkblue, linkcolor=darkblue, urlcolor=darkblue}
  \def\href#1#2{{#2}}
  \usepackage{url}
\title{Enforcing Consistency and Fairness in Multi-level Hierarchical Classification with a Mask-based Output Layer}

\author{
  Shijing Chen\textsuperscript{1},
  Shoaib Jameel\textsuperscript{2},
  Mohamed Reda Bouadjenek\textsuperscript{3},
  Feilong Tang\textsuperscript{4}, \\
    \textbf{Usman Naseem}\textsuperscript{5},
  \textbf{Basem Suleiman}\textsuperscript{1},
    \textbf{Hakim Hacid}\textsuperscript{6},
  \textbf{Flora D. Salim}\textsuperscript{1},
  \textbf{Imran Razzak}\textsuperscript{4,1} \\
  \textsuperscript{1}University of New South Wales, Sydney, NSW, Australia \quad
  \textsuperscript{2}University of Southampton, UK \\
  \textsuperscript{3}Deakin University, Australia \quad
  \textsuperscript{4}Mohamed Bin Zayed University of AI, UAE \\
  \textsuperscript{5}Macquarie University, Australia \quad
  \textsuperscript{6}Technology Innovation Institute, UAE \\
  \texttt{\{arthur.chen,hao.xue1,b.suleiman, flora.salim\}@unsw.edu.au }, \texttt{hakim.hacid@tii.ae} \\
  \texttt{m.s.jameel@southampton.ac.uk, \{feilong.tang, imran.razzak\}@mbzuai.ac.ae} \\ \\ \\ 
}

\begin{document}
\maketitle

\begin{abstract}
Traditional Multi-level Hierarchical Classification (MLHC) classifiers often rely on backbone models with $n$ independent output layers.
This structure tends to overlook the hierarchical relationships between classes, leading to inconsistent predictions that violate the underlying taxonomy. 
Additionally, once a backbone architecture for an MLHC classifier is selected, adapting the model to accommodate new tasks can be challenging.
For example, incorporating fairness to protect sensitive attributes within a hierarchical classifier necessitates complex adjustments to maintain the class hierarchy while enforcing fairness constraints. 
In this paper, we extend this concept to hierarchical classification by introducing a fair, model-agnostic layer designed to enforce taxonomy and optimize specific objectives, including consistency, fairness, and exact match.
Our evaluations demonstrate that the proposed layer not only improves the fairness of predictions but also enforces the taxonomy, resulting in consistent predictions and superior performance. 
Compared to Large Language Models (LLMs) employing in-processing de-biasing techniques and models without any bias correction, our approach achieves better outcomes in both fairness and accuracy, making it particularly valuable in sectors like e-commerce, healthcare, and education, where predictive reliability is crucial.
\end{abstract}

\section{Introduction}
The growing complexity of real-world datasets has led to the widespread use of multi-level hierarchical structures, making Multi-level Hierarchical Classification (MLHC) essential in modern data analysis. In domains such as e-commerce, where large-scale product datasets need effective categorization, MLHC plays a pivotal role \cite{silla2011survey,tieppo2022hierarchical}. For example, in an online store for beauty products, items are organized into a hierarchical taxonomy. At the top level (\(\ell_1\)) might be a broad category such as \textit{Beauty}, which branches into subcategories like \textit{Hair Care} at level (\(\ell_2\)), and even more specific classes like \textit{Hair Color} or \textit{Shampoo} at level (\(\ell_3\)). MLHC leverages these taxonomies to accurately classify items so that the results can be used to aid the recommendation of products based on their hierarchical relationships, which in turn enhances user experience, improves personalization, and drives sales. This hierarchical structure enables MLHC to capture semantic relationships between categories, making it indispensable in sectors like e-commerce where effective data organization and classification are crucial for scaling user interaction \cite{dumais2000hierarchical, agrawal2013multi,li2020hierarchical,shen2012large}.


\begin{figure*}[t]
    \centering    
    \subfloat[Amazon product review taxonomy.]{%
        \includegraphics[width=0.33\textwidth]{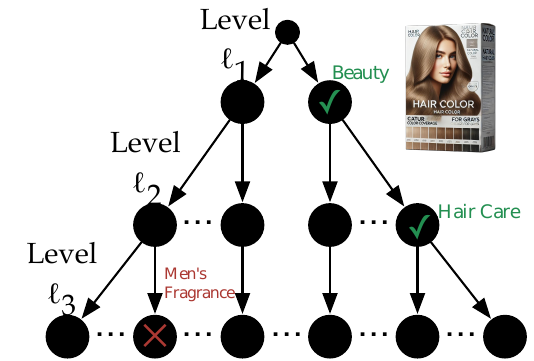}%
        \label{fig:hair}%
    }
    \subfloat[Product review taxonomy performance.]{%
        \includegraphics[width=0.34\textwidth]{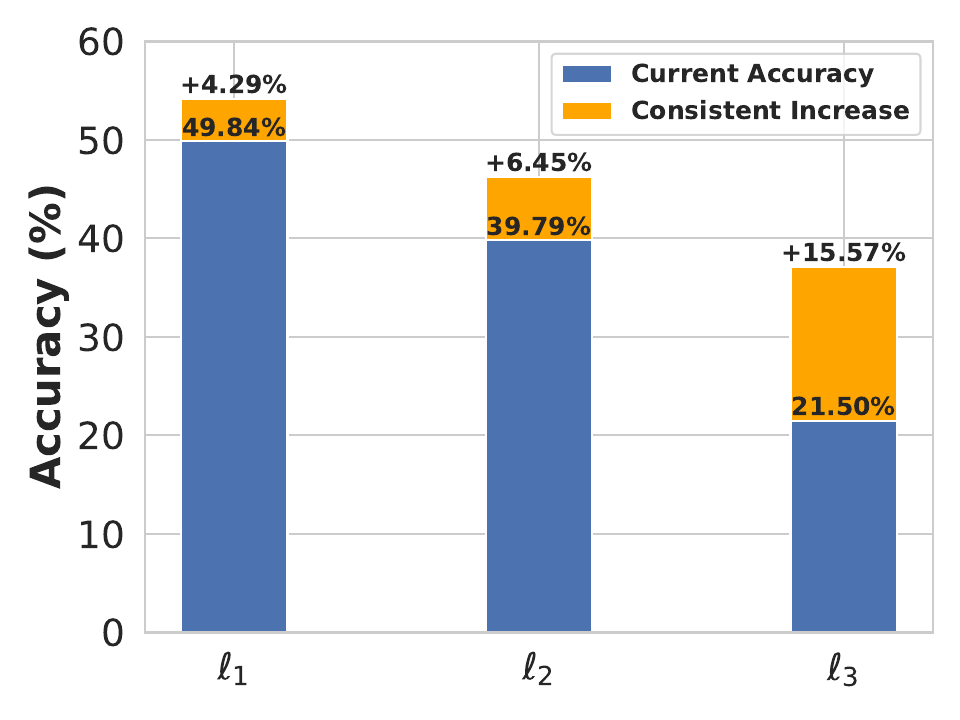}%
        \label{fig:potential}%
    }
    \subfloat[Gender performance difference.]{%
        \includegraphics[width=0.27\textwidth]{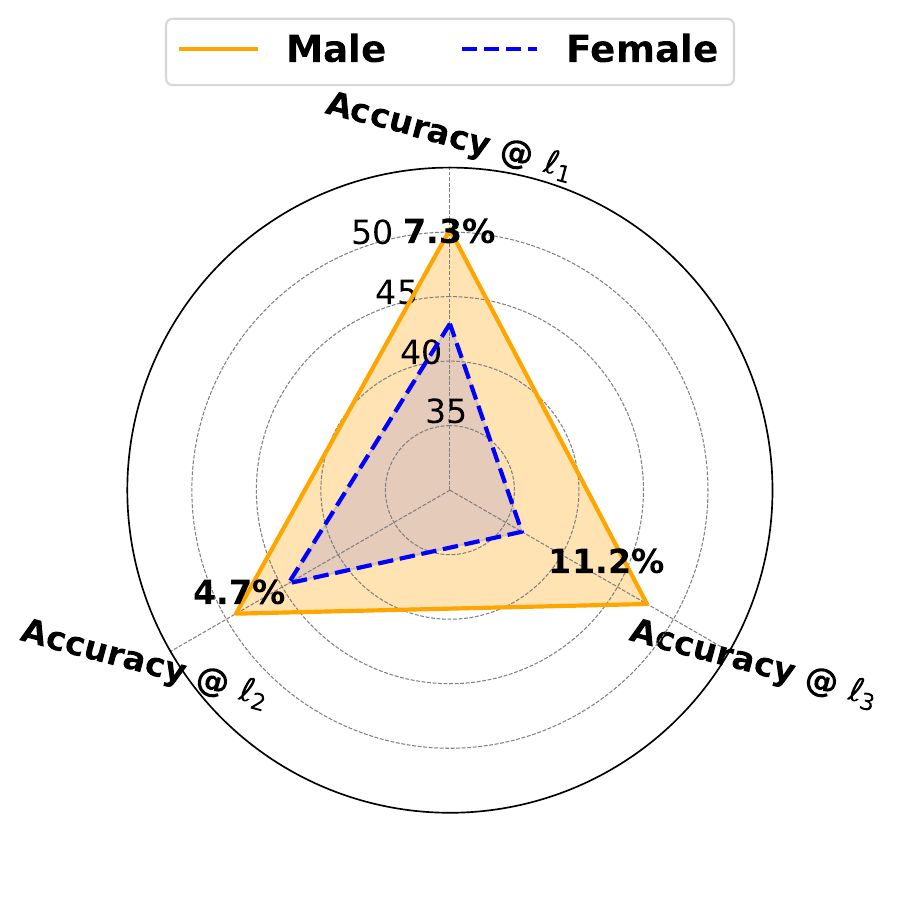}%
        \label{fig:gen}%
    }
    \caption{(a) Example of an Amazon product review classified across multiple levels of the taxonomy. (b) Proportion of correctly classified product reviews for each level of our taxonomy of the Amazon product review dataset, and the proportion of reviews incorrectly classified but for which other levels in the taxonomy were correctly identified. (c) Performance difference between male and female predictions using the BERT + Flat classifier model on the Amazon product review dataset. The percentages highlighted are actual accuracy differences between different genders.}
    \label{fig:combined}
\end{figure*}

Despite the advantages of MLHC, conventional methods still face significant challenges, particularly in ensuring both consistency and fairness across multiple levels of the hierarchy. Flat classifiers, which ignore the hierarchical relationships between categories, often result in inconsistent predictions, as shown in Figure \ref{fig:hair}. Figure \ref{fig:potential} demonstrates the potential accuracy gains that can be achieved by employing consistent hierarchical classifiers. For each level's accuracy shown in Figure \ref{fig:potential}, a portion of misclassified instances (4.29\% for $\ell_1$, 6.45\% for $\ell_2$, and 15.57\% for $\ell_3$) can be attributed to incorrect predictions while other levels in the hierarchy were correctly classified. By leveraging information from higher levels (\(\ell_1\) and \(\ell_2\)) that are correctly classified, the classifier can make more informed predictions at the lower levels (\(\ell_3\)), thus reducing misclassification. This suggests that accuracy at each level could improve if classifiers leverage the correct classifications from other levels in the taxonomy. Hence, enforcing consistency across hierarchical levels becomes a crucial and intriguing challenge in this domain. Additionally, existing models often inherit biases from the underlying data \cite{de2019bias, guo2021detecting, nangia2020crows}. Especially, Large Language Models (LLMs), which are trained on vast amounts of textual data, can amplify these biases, resulting in outputs that reinforce harmful stereotypes or exclude minority perspectives \cite{bender2021dangers, bommasani2021opportunities}. This highlights significant limitations of LLMs related to unfairness and bias, emphasizing the need for strategies to detect and mitigate these issues before utilizing their representational power. In large-scale datasets, unfairness and bias (e.g., age and gender) can significantly impact the fairness of predictions, resulting in unequal experiences for different users. Such biases undermine the reliability of machine learning systems and create ethical concerns, especially in applications where fairness is vital. Ensuring consistency in hierarchical classification is not only essential for improving overall accuracy but also plays a crucial role in addressing fairness. As shown in Figure \ref{fig:gen}, male predictions tend to be more accurate than female predictions across all levels of the hierarchy. This disparity highlights the need for a classifier that not only enforces consistency but also promotes fairness by ensuring that correct information at higher levels of the taxonomy is propagated downwards, improving performance for all demographic groups.

Different methods have been proposed for MLHC, which can be classified based on how they utilize the hierarchical structure. 
Specifically, we distinguish between three primary approaches: (i) the flat classification approach, where the class hierarchy is completely ignored. 
In this approach, predictions are made solely for the bottom levels, with the assumption that all ancestor classes are implicitly attributed to the instance as well; 
(ii) the local classification approach, which involves training a separate multi-class classifier at each parent node in the hierarchy to distinguish between its child nodes; 
and (iii) the global classification approach \cite{zhang2024teleclass, bettouche2024contextual, liu2024improve,chen2025leveraging}, where a single classifier is responsible for handling the entire class hierarchy.
In this paper, we argue that \textit{flat classifiers}, by ignoring the hierarchical relationships between class levels, often results in inconsistent classifications. 
For instance, as shown in Figure \ref{fig:combined}, the data entry of a \textit{Hair Color} product is correctly classified as \textit{Beauty} and \textit{Hair Care}, but incorrectly as \textit{Men's Fragrance} at the leaf node.
Furthermore, we argue that it is impractical to train and maintain $n$ separate networks for \textit{local classification approaches}, which can be redundant and costly in real-world applications. 
As a result, we favor \textit{global classification approaches}, which addresses the limitations of flat and local methods. 
However, existing methods still face several key challenges: 
(i) they do not inherently embed  the taxonomy structure, 
(ii) they often rely on complex neural network architectures with $n$ independent output layers that do not interact, 
(iii) they frequently produce predictions that are inconsistent with the taxonomy, 
and (iv) they typically operate with a fixed $n$, limiting flexibility and requiring extensive hyperparameter tuning to optimize $n$ for different scenarios.

Like traditional classifiers, MLHC models also inherit biases from the underlying data, potentially leading to unfair treatment of individuals based on protected characteristics such as race or gender. To address these challenges, we introduce a novel \textit{\textbf{Debiased Transitional Taxonomy Classifier (D-TTC)}}. Our approach features an LLM-agnostic output layer that integrates taxonomic information with a dynamic reweighting scheme to ensure fairness and balanced representation across demographic groups. Our D-TTC employs a \textit{top-down divide-and-conquer strategy}, attending to taxonomy relationships and applying fairness reweighting at each level of the hierarchy by broadcasting fairness and consistency from parents to children. This ensures that predictions remain consistent with the hierarchical structure while reducing biases. Unlike traditional methods that focus solely on accuracy, our model adjusts sample weights based on demographic factors like gender and race, promoting fairness throughout the classification process. We evaluate the effectiveness of our approach using the Amazon product review dataset and DBPedia dataset, leveraging various large language models as backbone classifiers. Experimental results demonstrate that D-TTC not only significantly reduces demographic biases but also improves hierarchical consistency and exact match rates, making it particularly valuable in sectors such as e-commerce, healthcare, and education where consistency, fairness, and predictive reliability are crucial.

\section{Related Work}

MLHC has been extensively studied across various domains. We review the most prominent approaches below.
\textbf{Flat and Local Classifier Approaches} ignore the hierarchical structure, predicting only leaf-node classes and implicitly assigning ancestor classes. While simple and efficient, they fail to leverage class relationships, leading to suboptimal performance in complex taxonomies~\cite{silla2011survey, valentini2010true}. To address these limitations, local classifiers approach train classifiers at different hierarchy levels. The \textit{Local Classifier per Node} (LCN) trains a classifier for each node~\cite{koller1997hierarchically} but can result in inconsistencies across levels~\cite{silla2011survey, dumais2000hierarchical}. The \textit{Local Classifier per Parent Node} (LCPN) trains classifiers for each parent node to distinguish among its children, reducing inconsistencies but potentially propagating errors down the hierarchy~\cite{secker2007experimental}. The \textit{Local Classifier per Level} (LCL), though less common, involves training classifiers at each level but may struggle with a large number of classes at deeper levels~\cite{de2009tutorial, costa2007comparing}.

\textbf{Global Approaches} treat the entire hierarchy as a single unit during training, integrating hierarchical information to ensure consistency across levels. Notable examples include the \textit{Clus-HMC} algorithm, which uses predictive clustering trees~\cite{kiritchenko2005functional, vens2008decision}. While these methods avoid error propagation inherent in local approaches, they require significant computational resources and often lack modularity~\cite{vens2008decision, silla2011survey}.

Extending global approaches, \textbf{Graph Neural Networks} (GNNs) model hierarchies as graphs with nodes representing labels and edges representing relationships, effectively capturing complex dependencies. Models like \textit{Hierarchy-Aware Graph Models} (HiAGM) have demonstrated improved performance across multiple levels~\cite{liu2023enhancing}. Additionally, specialized loss functions have emerged to ensure consistency in hierarchical multi-label classification. By incorporating a max constraint loss (MCLoss) that enforces hierarchical dependencies during training, methods like \textit{Coherent Hierarchical Multi-Label Classification Networks} (C-HMCNN) ensure coherent predictions where a child node is activated only if its parent node is~\cite{giunchiglia2020coherent}. This maintains logical consistency across hierarchical levels and significantly improves accuracy in domains where adherence to the hierarchy is critical. LLMs can also be utilized to enhance the performance for MLHC. TELEClass \cite{zhang2024teleclass}, which is proposed as a weakly-supervised MLHC framework, has employed a weakly-supervised approach by enriching label taxonomies with class-indicative terms using large language models (LLMs) and corpus-based analysis. This significantly improves pseudo-label quality and handles fine-grained classes, outperforming previous weakly-supervised and zero-shot LLM-based methods.

\textbf{Fairness} in machine learning is typically divided into two categories: \textit{individual fairness} and \textit{group fairness}. Individual fairness, such as counterfactual fairness \cite{kusner2017counterfactual}, ensures that a model provides similar outcomes for individuals who have similar attributes (e.g., age or race). In contrast, group fairness, like statistical parity \cite{dwork2012fairness}, assesses fairness across entire groups with the same protected attributes rather than focusing on individuals. This approach aims to ensure equitable treatment across different demographic cohorts. While both individual and group fairness addresses key aspects of fairness in machine learning, achieving these objectives in practice often requires mitigating bias within the models themselves. Bias can originate from various stages of model development, particularly in pretrained models, which can propagate bias to downstream tasks. Recent work has focused on mitigating intrinsic bias during pretraining and in-processing stages, using various techniques such as \textit{Counterfactual Data Augmentation (CDA)}, \textit{Context-debias}, and \textit{Sent-debias}. For instance, \textit{CDA} balances representation by swapping demographic-specific terms (e.g., "he" and "she") in the training data, though it is resource-intensive due to the need for retraining \cite{zmigrod2019counterfactual, webster2020measuring}. In-processing methods like \textit{Context-debias} attempt to remove bias by ensuring that embeddings of stereotypical terms are orthogonal to gender-related terms, but they depend heavily on predefined word lists, limiting their generalizability \cite{kaneko2021debiasing}. Post-processing methods such as \textit{Sent-debias} work by removing gender bias from pretrained model embeddings, though research suggests that these methods often obscure rather than fully eliminate bias \cite{liang2020towards, gonen2019lipstick}. 

Despite significant advancements in integrating deep learning techniques for tasks involving multi-level taxonomies, challenges persist in scaling models to handle large, complex hierarchies consistently and fairly. Existing methods often struggle to maintain consistency across deep hierarchies. Also, upstream debiasing techniques frequently do not translate into improved fairness in downstream tasks—especially in complex scenarios like MLHC where preserving class hierarchy is crucial \cite{steed2022upstream}. To tackle these issues, we propose the \textit{Debiased Taxonomy-based Transitional Classifier} (D-TTC), which embeds hierarchical information directly into the classification process, leverages LLMs for better contextual understanding, and uses downstream post-processing debiasing through dynamic reweighting which adjusts the importance of different samples during training.  As a model-agnostic layer, D-TTC enhances both flexibility and performance across various backbone models, providing a more consistent solution for complex hierarchies. Additionally, it enables the model to address bias more effectively within specific application domains, ensuring improved fairness alongside high performance.

\section{Notations and problem definition}

Generally, the classification problems are flat classification, where each input instance is assigned to a single output class from a finite set of independent, non-hierarchical classes. Formally, given a dataset $\mathcal{D} = \{(\textbf{x}^{(1)},y^{(1)}), (\textbf{x}^{(2)},y^{(2)}),$ $\cdots, (\textbf{x}^{(m)},y^{(m)})\}$ with $m$ instances, where each $\textbf{x}^{(i)}\in \mathbb{X} \subseteq \mathbb{R}^n$ is an $n$-dimensional input feature vector of the instance $i$ and $y^{(i)}\in \mathcal{Y}=\{y_1,y_2,\cdots,y_k\}$ represents its class, 
a classification algorithm must learn a mapping function $f:\mathbb{X} \rightarrow \mathcal{Y}$, which maps each feature vector $\textbf{x}^{(i)}$ to its corresponding class $y^{(i)}$. However, unlike \textit{flat classification} where the classes are considered unrelated, in a hierarchical classification, classes are structured in a taxonomy, which is typically structured as a tree, where each class has one parent or as a directed acyclic graph (DAG), where a class may have several parents.
Given a set of classes $\mathcal{Y}$, Wu et al.~\cite{wu2005learning} defined a taxonomy as a pair $(\mathcal{Y},  \prec)$, where $\prec$ is the \textit{``subclass-of''} relationship with the following properties~\cite{wu2005learning,silla2011survey}: 
(i) asymmetry ($\forall y_i, y_j \in \mathcal{Y}, if y_i \prec y_j$ then $y_j \nprec y_i$), 
(ii) anti-reflexivity ($\forall y_i \in \mathcal{Y}, y_i \nprec y_i)$, 
and (iii) transitivity ($\forall y_i, y_j, y_k \in \mathcal{Y}, y_i \prec y_j$ and $y_j \prec y_k$ implies $y_i \prec y_k$).

In hierarchical classification, \textit{fairness} refers to the equitable treatment of instances from different demographic groups. Let \( \mathcal{G} = \{g_1, g_2, \cdots, g_q\} \) represent the set of demographic groups (e.g., gender, race), and each instance \( \textbf{x}^{(i)} \in \mathcal{D} \) is associated with a group label \( g^{(i)} \in \mathcal{G} \). A classifier is considered fair if the probability of correct classification is independent of the demographic group \( g \), i.e., the performance of the classifier should not systematically favor or disadvantage any subgroup.  For a hierarchical classification model \( f \), fairness can be expressed as: $\scalemath{0.76}{ \mathbb{P}\left( f(\textbf{x}^{(j)}) = y_{[\ell_i]}^{(j)} \mid g^{(j)} = g_n \right)  = \mathbb{P}\left( f(\textbf{x}^{(j)}) = y_{[\ell_i]}^{(j)} \mid g^{(j)} = g_m \right)}$. The above equation states that the probability of correct classification for any data $j$ should be equal across all demographic groups at any given hierarchical level \( \ell_i \). In contrast, \textit{bias} refers to the systematic difference in the classifier's performance for different demographic groups. Thus can be defined as the deviation in classification accuracy for group \( g \) relative to the overall accuracy across all groups: \scalemath{0.76}{ \text{Bias}_{g, \ell_i} = \mathbb{P}(f(\textbf{x}^{(j)}) = y_{[\ell_i]}^{(j)} \mid g^{(j)} = g) - \mathbb{P}(f(\textbf{x}^{(j)}) = y_{[\ell_i]}^{(j)})}. A classifier is unbiased if \( \text{Bias}_{g, \ell_i} = 0 \) for all \( g \in \mathcal{G} \) and \( \ell_i \). Any deviation from zero indicates that the classifier is biased toward or against certain demographic groups at that hierarchical level. However, in binary classification, this requires that the true positive rate (TPR) and false positive rate (FPR) are the same across all groups: 
$\scalemath{0.76}{ \text{TPR}_g = \mathbb{P}(f(\textbf{x}^{(i)}) = 1 \mid y^{(i)} = 1, g^{(i)} = g)}$, and 
$\scalemath{0.76}{ \text{FPR}_g = \mathbb{P}(f(\textbf{x}^{(i)}) = 1 \mid y^{(i)} = 0, g^{(i)} = g)}$. 
Equalized Odds ensures fairness by requiring that $\text{TPR}_g$ and $\text{FPR}_g$ are consistent across all demographic groups $g \in \mathcal{G}$, meaning that the model’s performance is independent of group membership. The bias can be measured by the deviation between $\text{TPR}_g$ and $\text{FPR}_g$ for different demographic groups. $\scalemath{0.76}{\text{Bias}_{g, \ell_i}=\max(| \text{TPR}_{g_n}-\text{TPR}_{g_m}|, | \text{FPR}_{g_n}-\text{FPR}_{g_m}|)
}$. Where $g_m$ and $g_n$ represent different demographic groups.

We measure fairness using \textit{Equalized Odds} \cite{hardt2016equality}, which ensures that the classifier's prediction is independent of the demographic group \( g^{(i)} \in \mathcal{G} \), conditioned on the true label. Specifically, for any demographic group \( g^{(i)} \), Equalized Odds require that the true positive rate (TPR) and false positive rate (FPR) are equal across all groups. Formally, Equalized Odds is satisfied when: \scalemath{0.76}{
\mathbb{P}(f(\textbf{x}^{(i)}) = \hat{y}^{(i)} \mid y^{(i)} = y, g^{(i)} = g) = \mathbb{P}(f(\textbf{x}^{(i)}) = \hat{y}^{(i)} \mid y^{(i)} = y)
} for all \( g \in \mathcal{G} \), meaning that the model's predictions are conditionally independent of the demographic group.

\noindent\textbf{Problem definition:} In this study, we focus on \textit{tree} taxonomies, which follow a hierarchical structure with $n$ levels \(\ell_i\). These levels satisfy the conditions \(\ell_i \subset \mathcal{Y}\) and \(\ell_1 \cup \ell_2 \cup \cdots \cup \ell_n = \mathcal{Y}\). For all \(y_j \in \ell_1\), \(y_j\) has no parent, and for every \(y_j \in \ell_{i+1}\), there exists exactly one \(y_k \in \ell_{i}\) such that \(y_j\) is a descendant of \(y_k\) for \(i \ge 1\) (see Figure \ref{fig:hair} for an example of a three-level taxonomy).  We represent the relationship between two consecutive levels \(\ell_i\) and \(\ell_{i+1}\) using an \(|\ell_i| \times |\ell_{i+1}|\) matrix \(M^{[\ell_i,\ell_{i+1}]}\), where the binary value \(M^{[\ell_i,\ell_{i+1}]}_{y_k,y_j} \in \{0 \ (\text{if } y_j\) is not a descendant of \( y_k), 1 \ (\text{if } y_j \text{ is a descendant of } y_k)\}\), with \(y_k \in \ell_i\) and \(y_j \in \ell_{i+1}\). The multi-level hierarchical classification problem addressed is defined as the task of learning a mapping function $f:(\mathbb{X}_1 \times \mathbb{X}_2 \times \cdots \times \mathbb{X}_p) \rightarrow \mathcal{Y}$ which assigns to each instance--represented by a combination of feature vectors from $p$ different modalities--a prediction vector $\textbf{y}^{(i)} = \{y^{[\ell_1]}, y^{[\ell_2]},\cdots, y^{[\ell_n]}\}$. Here, $y^{[\ell_i]} \in \ell_i$ represents the class assigned by the function $f$ at each hierarchical level $\ell_i$, ensuring not only accurate and consistent predictions but also better bias mitigation across all taxonomy levels. 

\section{Fair Model Agnostic Hierarchical Framework}

\begin{figure}[!t]
    \centering
    \includegraphics[height=4.1cm, width=0.99\linewidth]{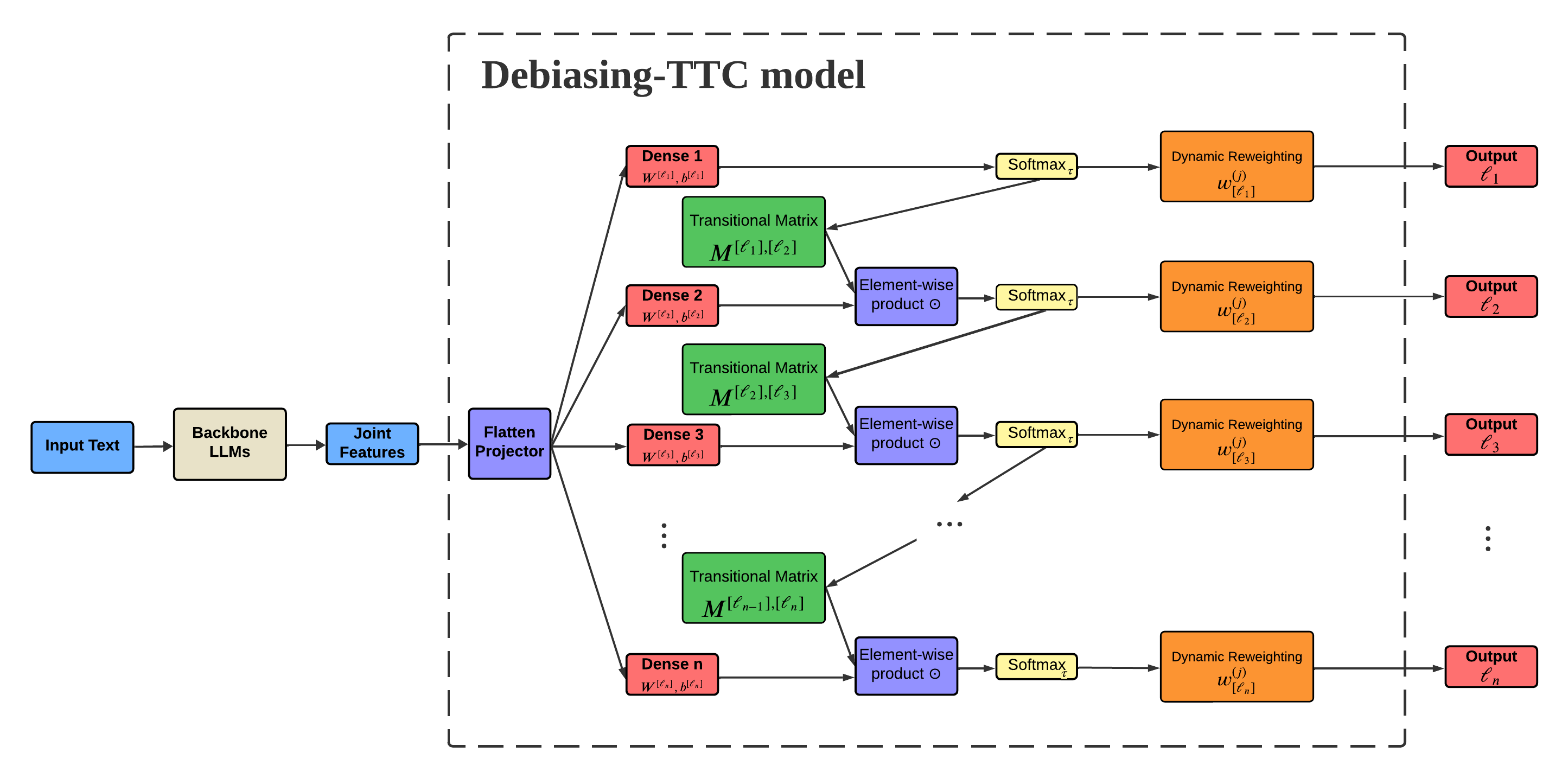}
        \vspace{-10pt}
    \caption{Architecture for Debiased-TTC model layers.}
    \label{fig:arch}
        \vspace{-10pt}
\end{figure}

We extend the concept of fairness to hierarchical classification by incorporating de-biasing factors and taxonomy into a model-agnostic layer specifically designed to enforce the hierarchical structure while optimizing key objectives such as consistency, fairness, and exact match accuracy. Our approach ensures that predictions not only respect the taxonomy but also improve fairness across different categories. Following, we first introduce taxonomy-based transitional classifier (TTC), followed by the integration of debiasing at each hierarchical level. These components propagate fairness and consistency from parent nodes to child nodes, ensuring a balanced and structured prediction process.

\subsection{TTC Model Description}

 We present the taxonomy-based transitional classifier which overcomes the aforementioned shortcomings of existing methods, which often lead to contradictory predictions, by ensuring consistency at every stage of the prediction process. The TTC layer utilizes the detailed taxonomy at each hierarchical level to constrain its predictions to valid labels for the respective level. This helps prevent misclassifications across unrelated categories. By embedding the hierarchy directly into the model, the TTC layer promotes coherence in predictions and seeks to improve accuracy in text data, potentially surpassing traditional classifiers.

Figure \ref{fig:arch} illustrates the architecture of the proposed TTC layer, an LLM-agnostic component designed to leverage the taxonomy and ensure that predictions adhere to the hierarchical structure of the data.
Several independent classifiers are used to predict the categories on different levels in the same way as local approaches. 
However, to maintain consistency, the relation information of upper levels is incorporated into the next level in the same way as attention is. 
The output probabilities from the upper level are multiplied by a \textit{transition} matrix, where each entry represents the relationship between classes at successive levels in the taxonomy (i.e., 1 if the class in the column is a ``subclass of'' the class in the row, and 0 otherwise).
The product can be considered as the attention score that incorporates the hierarchical information as well as the relation between classes and can be applied to the output probability for the next level. 
The prediction of the classifiers can be formulated as $\textbf{z}^{[\ell_i]} = W^{[\ell_i]} \cdot \textbf{a} + b^{[\ell_i]}$, where $\textbf{a}$ is the joint output latent feature of backbone LLMs, and $W^{[\ell_i]}, b^{[\ell_i]}$ are learnable parameters that trained on the trainset regarding each $\ell_i$ of the hierarchies. 
The prediction of the first classifier is obtained by applying a temperature-scaled \textit{softmax} normalization, as $\hat{\textbf{y}}^{[\ell_1]} = \text{softmax}(\textbf{z}^{[\ell_1]})$. 
For each subsequent level, we compute an attention score to incorporate relational information into the predictions, ensuring consistency across levels (i.e., $\hat{y}^{[\ell_{i+1}]} \prec \hat{y}^{[\ell_i]}$). This is achieved by injecting hierarchical relations as follows:
\begin{equation} \label{equation-attention}
\textbf{m}^{[\ell_{i+1}]} = \hat{\textbf{y}}^{[\ell_i]} \times M^{[\ell_i, \ell_{i+1}]}
\end{equation}

\noindent where \(M^{[\ell_i, \ell_{i+1}]}\) is our \( |\ell_i| \times |\ell_{i+1}| \) transitional matrix which encodes the relationship between two successive levels \( \ell_i \) and \( \ell_{i+1} \) in a taxonomy (i.e., the binary value \( M^{[\ell_i, \ell_{i+1}]}_{y_k,y_j} \in \{ 0 \ (\text{if } y_j \not\prec y_k), 1 \ (\text{if } y_j \prec y_k) \} \), with \( y_k \in \ell_i \) and \( y_j \in \ell_{i+1} \)). 
Referring to the example illustrated in Figure \ref{fig:hair}, consider the \(\ell_2\) labels, which include \textit{Hair Care} and \textit{Cosmetics}, and the \(\ell_3\) labels, comprising \textit{Hair Color}, \textit{Shampoo}, \textit{Lipsticks}, and \textit{Skin Care}. 
The corresponding transition matrix \(M^{[\ell_2, \ell_3]}\) is: 
\[ 
    M^{[\ell_2, \ell_3]} = 
    \begin{pmatrix}
        1 & 1 & 0 & 0 \\
        0 & 0 & 1 & 1
    \end{pmatrix}
\]

\noindent in which the first row corresponds to the  \(\ell_2\)  class \textit{Hair Care}, where a value of 1 indicates that the  \(\ell_3\)   class (e.g., \textit{Hair Color} or \textit{Shampoo}) is a subclass of \textit{Hair Care}, and a value of 0 indicates no such relationship. 
Similarly, the second row refers to the \(\ell_2\) class \textit{Cosmetics}, where the values reflect whether the \(\ell_3\) classes are subclasses of \textit{Cosmetics}. 
In this manner, the hierarchical structure of the taxonomy is fully encapsulated within the transitional matrix $M$. Each attention score is applied using an element-wise product on the probability output of each classifier from a lower level  as:
\begin{equation} \label{equation-predictions}
    \hat{\textbf{y}}^{[\ell_{i+1}]} = \textit{softmax}_{\tau} (\textbf{z}^{[\ell_{i+1}]} \circ \textbf{m}^{[\ell_{i+1}]})
\end{equation}

Attention scores and classifications in Equations~\ref{equation-attention} and \ref{equation-predictions}, respectively, are processed sequentially for all hierarchical levels. The loss function is also adjusted as follows:

\begin{equation}
\frac{1}{m} \sum_{j=1}^{m} \sum_{i=1}^{n} \left[ \pi^{[\ell_i]} \cdot \mathcal{L}(y^{(j)}_{[\ell_i]}, \hat{y}^{(j)}_{[\ell_i]}) \right]
\end{equation}

where $\mathcal{L} (\bullet,\bullet )$ denotes the cross-entropy function and $\pi^{[\ell_i]}$ are a set of importance factors that can be tuned to changing the weight of losses for different \(\ell_i\). 

Continuing with the example provided earlier, given the \textit{transition} matrix \(M^{[\ell_2, \ell_3]}\), and assuming the probability output from the \(\ell_2\) classifier is \(\hat{\textbf{y}}^{[\ell_2]} = \{0.9, 0.1\}\), the attention scores are calculated as: 
\(
\textbf{m}^{[\ell_3]} = \hat{\textbf{y}}^{[\ell_2]} \cdot M^{[\ell_2, \ell_3]} = \{0.9, 0.9, 0.1, 0.1\}
\).
Assuming the output from \(\ell_3\) is \(\textbf{z}^{[\ell_3]} = \{-0.2, 0.5, 1.3, 0.3\}\), applying the attention scores \(\textbf{m}^{[\ell_3]}\) and a softmax function to normalize the result gives the prediction probability output:
\(
\hat{\textbf{y}}^{[\ell_3]} = \{0.182, 0.342, 0.249, 0.225\}.
\)

Compared to a flat classifier for \(\ell_3\) which would have applied directly \textit{softmax} to \(\textbf{z}^{[\ell_3]}\),  TTC's prediction produces more consistency with upper-level prediction. 
Additionally, from a taxonomic perspective, \textit{tree-like} hierarchical classification leverages general-to-specific relationships, where general categories have better data separability. This indicates that they possess wider margins in their decision boundaries, making it easier for classifiers to distinguish them. As a result, general classes at higher levels contribute to higher classification accuracy at the top \cite{cortes1995support}. By enforcing consistency across hierarchical levels, the LLM is further guided to make more accurate predictions at deeper, more specific levels with greater granularity.

\subsection{Fairness in TTC}

While the TTC was initially designed as a model-agnostic layer to ensure consistency across hierarchical levels, it is also important to address potential fairness concerns. Specifically, in scenarios where certain demographic groups, such as gender, are over-represented or under-represented, bias can arise. To mitigate this, we introduce a dynamic reweighting mechanism within the TTC framework to promote fairer predictions. The dynamic reweighting scheme adjusts the weight of each sample based on its demographic group, such as gender. The weight assigned to each sample at a given level \( \ell_i \) is defined as: 
\[
w_{\ell_i}^{(j)} = 
\begin{cases} 
\frac{1}{N_{g,\ell_i} + \epsilon} & \text{if } d_j = G \\ 
1 & \text{if } d_j = N 
\end{cases}
\]
where \( N_{g,\ell_i} \) represents the count of samples from a particular demographic group \( g \) at level \( \ell_i \), and \( d_j \) indicates the group membership of the sample (e.g., Female, Male). The small constant \( \epsilon \) is included to avoid division by zero.

The count \( N_{g,\ell_i} \) is calculated as: $\scalemath{0.7}{N_{g, \ell_i} = \sum_{j=1}^{m} \mathbb{I}(g_j = g) \cdot \mathbb{I}(\hat{y}_{\ell_i}^{(j)} = c)}$ where \( \mathbb{I}(\cdot) \) is the indicator function, \( g_j \) represents the demographic group of sample \( j \), and \( y_{\ell_i}^{(j)} \) is the predicted label at level \( \ell_i \). For example, if at level \( \ell_2 \) there are 30 samples labeled as \textit{Female} and predicted as \textit{Hair Care}, then the dynamic weight for these samples for this training iteration would be: $\scalemath{0.6}{w_{\ell_2}^{(j)} = \frac{1}{30 + \epsilon} \approx 0.03} $. 
This reweighting mechanism ensures that samples from under-represented groups contribute more heavily to the training process, thereby addressing potential imbalances in the dataset. More importantly, it also encourages the model to focus more effectively on neutral samples that are not affected by sensitive attributes. The dynamic weights are incorporated into the overall loss function as follows: $ \scalemath{0.6}{
L_{\text{weighted}} = \frac{1}{m} \sum_{j=1}^{m} \sum_{i=1}^{n} \left[ \pi^{[\ell_i]} \cdot w_{\ell_i}^{(j)} \cdot \mathcal{L}(y_{\ell_i}^{(j)}, \hat{y}_{\ell_i}^{(j)}) \right]}$. where \( \mathcal{L}(\cdot, \cdot) \) denotes the cross-entropy loss for the sample, and \( \pi^{[\ell_i]} \) are the importance factors that adjust the relative contribution of different hierarchical levels. By incorporating this dynamic reweighting mechanism, the TTC model is able to address fairness concerns, ensuring that predictions are not biased towards over-represented groups. This adjustment, combined with the hierarchical consistency provided by the TTC layer, allows for a fairer and more balanced classification outcome across all hierarchical levels. The approach is particularly effective in real-world applications where demographic bias must be minimized to ensure equitable results.


\section{Experiments}
In this section, we analyze the effectiveness of our proposed taxonomy classifier and impact of debasing at different hierarchical levels. To evaluate the performance, we have used two hierarchical datasets: \textit{Amazon product review} and \textit{DBPedia}. 
We employed seven pre-trained LLMs to extract features from the textual data. Following the feature extraction, we applied our D-TTC classifier to classify the reviews across all three hierarchical levels. The training process was optimized to ensure that the proposed framework leveraged the hierarchical structure while minimizing bias.


\subsection{Experimental Setup}

\begin{figure}[!htb]
    \centering
    \includegraphics[width=\columnwidth]{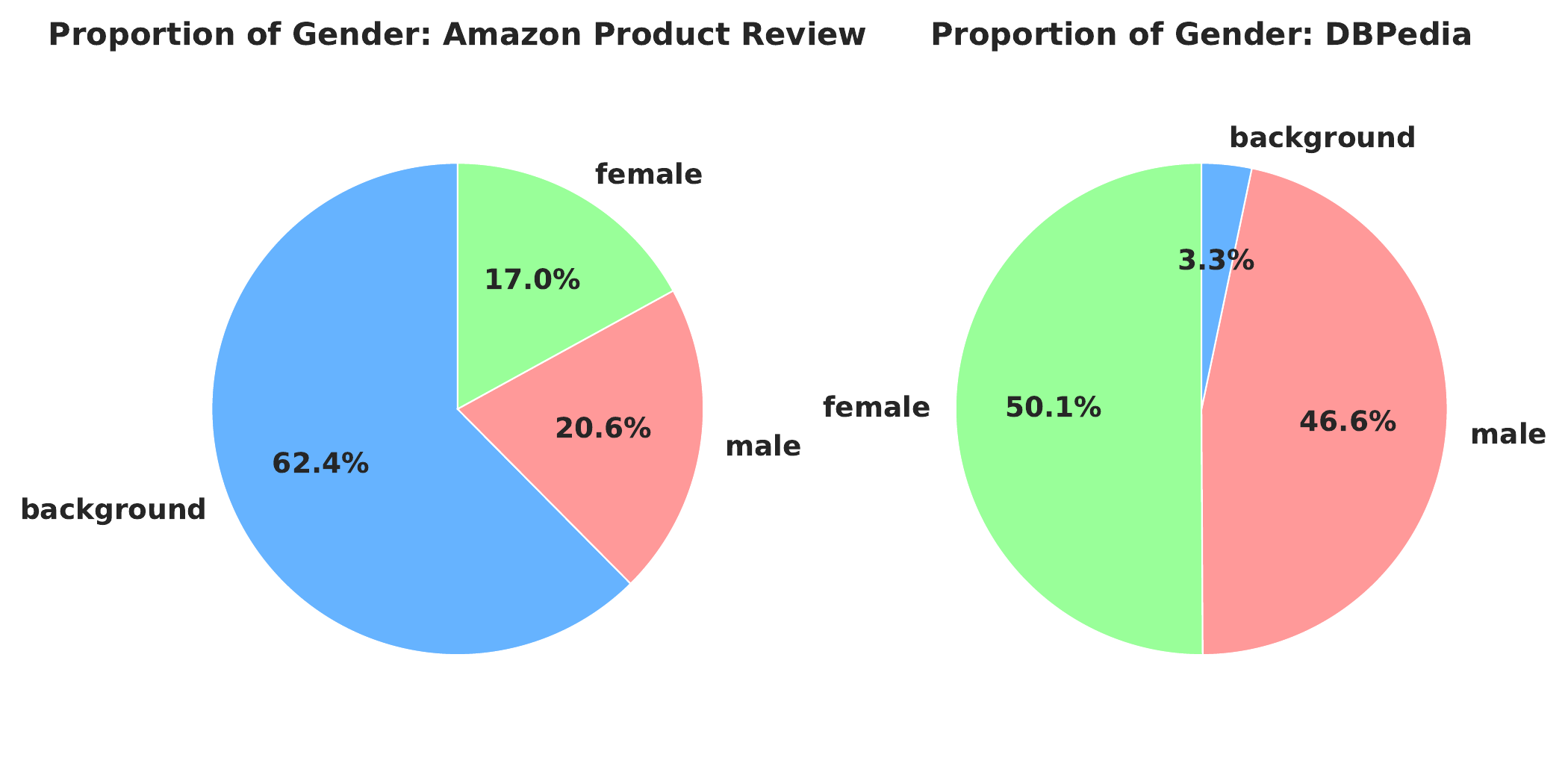}
        \vspace{-15pt}
    \caption{The gender distribution of two datasets.}
    \label{fig:gender}
    \vspace{-15pt}
\end{figure}

\noindent\textbf{Datasets:} 
The datasets that we have used are the \textit{Amazon Product Review} \cite{yury_kashnitsky_2020} and \textit{DBPedia} \cite{lehmann2015dbpedia}. The \textit{Amazon Product Review} dataset is large-scale, containing over 50,000 consumer reviews across various product categories. It includes structured data such as product IDs, review text, user ratings, helpfulness scores, and a three-level hierarchical classification system (with 6, 64, and 510 classes) that organizes products into broad categories (e.g., grocery, toys) and more specific subcategories, offering a detailed view of customer feedback and product classifications. \textit{DBPedia} is a large-scale dataset that provides structured, taxonomic, and hierarchical categories for over 90,000 Wikipedia articles across three levels (9, 70, and 219 classes), commonly used as a baseline for NLP and text classification tasks. We have applied gender-related keyword search, going through all dataset input to classify them as three subgroups: \textit{Male}, \textit{Female}, and \textit{Background}. The detailed distribution of the two datasets is shown in Figure \ref{fig:gender}. The gender distribution across the Amazon Product Review and DBPedia datasets highlights key differences. Amazon reviews are predominantly gender-neutral, reflecting a focus on products rather than individuals, with only a small proportion explicitly identifying male or female. In contrast, DBPedia shows a more balanced gender representation, as its entries primarily describe human entities, leading to more explicit gender markers. These distinctions underscore the differing content focus of each dataset, with Amazon being product-centric and DBPedia being entity-centric. The datasets were split to support robust model training and evaluation. The Amazon Product Review dataset was divided into 40,000 samples for training and 10,000 samples for testing, ensuring a substantial training set while reserving a portion for validation. In contrast, the DBPedia dataset had a larger split, with 60,000 samples used for training and 30,000 samples for testing. This larger test set allows for a more comprehensive evaluation of model performance, reflecting the dataset's entity-centric nature and the need for broader testing coverage.


\noindent\textbf{Backbone LLMs:} For backbone models, we have adopted different LLMs including: Bert\cite{devlin2018bert}, GPT-2\cite{radford2019language}, T5\cite{raffel2020exploring}, Qwen\cite{qwen2023}, Gemma\cite{gemma2023}, Phi3-mini\cite{phi3mini2024} and Llama 2 (7B) \cite{touvron2023llama}. We employed pre-trained LLMs with INT8 quantization to reduce memory usage and improve computational efficiency. The model extracted features from textual data by tokenizing input text context. Batching was used to manage memory, and the final hidden states were pooled with attention masks to generate feature vectors, which serve as a unified representation of the textual data, capturing relevant patterns across both datasets. By using attention masks to exclude padding tokens, the resulting latent features had greater representation power, as they focused on the meaningful parts of the input. These features were stored in compressed HDF5 format, enabling scalable processing for downstream tasks such as training and testing the subsequent classification modules. After feature extraction using the fine-tuned LLM, we applied a D-TTC classifier for hierarchical classification. The D-TTC model was designed to classify the reviews across all three levels of the hierarchy, ensuring consistency across the levels and minimizing bias during the classification process.


\begin{table*}[!htb]
\centering
\tiny
    \caption{Performance of Large Language Models with and without TTC on Amazon Product Review and DBPedia. The ablation studies were conducted by applying different modules independently. \textit{(D)} refers to flat classifiers with Dynamic Reweighting, \textit{(H)} represents TTC classifiers, and \textit{(HD))} denotes the D-TTC classifier}
    \label{tab:results}
    \resizebox{!}{122pt}{
\begin{tabular}{llrrrrrrrr}
\toprule
& & \multicolumn{4}{c}{Amazon Product Review} & \multicolumn{4}{c}{DBPedia} \\
\cmidrule(lr){3-6} \cmidrule(lr){7-10}
         Model & & HF1 &  Consistency &  Exact Match &  EO(Avg) & HF1 &  Consistency &  Exact Match &  EO(Avg) \\
\midrule
        Bert & & 0.3679 & 0.3278 & 0.1586 & 0.0772 & \textbf{0.6954} & 0.5832 & 0.3417 & 0.0663 \\
       
       Bert(D) & & 0.3699 \textcolor{ForestGreen}{(+0.0020)} & 0.3386 \textcolor{ForestGreen}{(+0.0108)} & 0.1522 \textcolor{red}{(-0.0064)} & 0.0755 \textcolor{ForestGreen}{(-0.0017)} & 0.6761 \textcolor{red}{(-0.0193)} & 0.5948 \textcolor{ForestGreen}{(+0.0116)} & 0.3265 \textcolor{red}{(-0.0152)} & 0.0577 \textcolor{ForestGreen}{(-0.0086)} \\
       
       Bert(H) & & 0.4288 \textcolor{ForestGreen}{(+0.0609)} & 0.3673 \textcolor{ForestGreen}{(+0.0395)} & 0.1680 \textcolor{ForestGreen}{(+0.0094)} & 0.0746 \textcolor{ForestGreen}{(-0.0026)} & 0.6897 \textcolor{red}{(-0.0057)} & 0.6243 \textcolor{ForestGreen}{(+0.0411)} & \textbf{0.3569 }\textcolor{ForestGreen}{(+0.0152)} & 0.0651 \textcolor{ForestGreen}{(-0.0012)} \\
      
      Bert(HD) & & \textbf{0.4346} \textcolor{ForestGreen}{(+0.0667)} & \textbf{0.3681} \textcolor{ForestGreen}{(+0.0403)} & \textbf{0.1687} \textcolor{ForestGreen}{(+0.0101)} & \textbf{0.0694} \textcolor{ForestGreen}{(-0.0078)} & 0.6725 \textcolor{red}{(-0.0229)} & \textbf{0.6412} \textcolor{ForestGreen}{(+0.0580)} & 0.3384 \textcolor{red}{(-0.0033)} & \textbf{0.0562} \textcolor{ForestGreen}{(-0.0101)} \\
\midrule
    Llama-2-7B & & 0.3996 & 0.3585 & 0.1626 & 0.0749 & 0.7023 & 0.6354 & 0.3492 & 0.0689 \\
 Llama-2-7B(D) & & 0.4030 \textcolor{ForestGreen}{(+0.0034)} & 0.3457 \textcolor{red}{(-0.0128)} & 0.1364 \textcolor{red}{(-0.0262)} & 0.0730 \textcolor{ForestGreen}{(-0.0019)} & 0.6505
\textcolor{red}{-0.0518} & 0.6211\textcolor{red}{-0.0143} & 0.3324\textcolor{red}{-0.0168} & \textbf{0.0568}\textcolor{ForestGreen}{-0.0121} \\
 
 Llama-2-7B(H) & & 0.4013 \textcolor{red}{(-0.0017)} & 0.4004 \textcolor{ForestGreen}{(+0.0419)} & 0.1717 \textcolor{ForestGreen}{(+0.0091)} & 0.0707 \textcolor{ForestGreen}{(-0.0042)} & \textbf{0.07108}\textcolor{ForestGreen}{+0.0085} & 0.6485\textcolor{ForestGreen}{+0.0131} & \textbf{0.3655}\textcolor{ForestGreen}{+0.0163} & 0.0666\textcolor{ForestGreen}{-0.0023} \\

Llama-2-7B(HD) & & \textbf{0.4203} \textcolor{ForestGreen}{(+0.0207)} & \textbf{0.4116} \textcolor{ForestGreen}{(+0.0531)} & \textbf{0.1723} \textcolor{ForestGreen}{(+0.0097)} & \textbf{0.0678} \textcolor{ForestGreen}{(-0.0071)} & 0.6872\textcolor{red}{-0.0151} & \textbf{0.6593}\textcolor{ForestGreen}{+0.0239} & 0.3411\textcolor{red}{-0.0081} & 0.0616\textcolor{ForestGreen}{-0.0073} \\

\midrule
    
    GPT-2 & & 0.3896 & 0.3157 & 0.1338 & 0.1110 & \textbf{0.6671} & 0.5729 & 0.3415 & 0.0777 \\
 
 GPT-2(D) & & \textbf{0.3992} \textcolor{ForestGreen}{(+0.0096)} & 0.3010 \textcolor{red}{(-0.0147)} & 0.1240 \textcolor{red}{(-0.0098)} & 0.0978 \textcolor{ForestGreen}{(-0.0132)} & 0.6647 \textcolor{red}{(-0.0024)} & 0.5810 \textcolor{ForestGreen}{(+0.0081)} & 0.3198 \textcolor{red}{(-0.0217)} & 0.0738 \textcolor{ForestGreen}{(-0.0039)} \\
 
 GPT-2(H) & & 0.3923 \textcolor{ForestGreen}{(+0.0027)} & \textbf{0.3351} \textcolor{ForestGreen}{(+0.0194)} & 0.1341 \textcolor{ForestGreen}{(+0.0003)} & 0.1137 \textcolor{red}{(+0.0027)} & 0.6489 \textcolor{red}{(-0.0182)} & 0.5962 \textcolor{ForestGreen}{(+0.0233)} & \textbf{0.3519} \textcolor{ForestGreen}{(+0.0104)} & 0.0729 \textcolor{ForestGreen}{(-0.0048)} \\

GPT-2(HD) & & 0.3868 \textcolor{red}{(-0.0028)} & 0.3235 \textcolor{ForestGreen}{(+0.0078)} & \textbf{0.1495} \textcolor{ForestGreen}{(+0.0157)} & \textbf{0.0954} \textcolor{ForestGreen}{(-0.0156)} & 0.6412 \textcolor{red}{(-0.0259)} & \textbf{0.6143 }\textcolor{ForestGreen}{(+0.0414)} & 0.3281 \textcolor{red}{(-0.0134)} & \textbf{0.0711} \textcolor{ForestGreen}{(-0.0066)} \\

\midrule
        T5 & & \textbf{0.4055} & 0.3604 & \textbf{0.1717} & 0.0853 & 0.7239 & 0.6114 & 0.3764 & 0.0597 \\
     
     T5(D) & & 0.3935 \textcolor{red}{(-0.0120)} & 0.3601 \textcolor{red}{(-0.0003)} & 0.1532 \textcolor{red}{(-0.0185)} & 0.0900\textcolor{ForestGreen}{(+0.0047)} & 0.7124 \textcolor{red}{(-0.0115)} & 0.6159 \textcolor{ForestGreen}{(+0.0045)} & 0.3687 \textcolor{red}{(-0.0077)} & 0.0577 \textcolor{ForestGreen}{(-0.0020)} \\
     
     T5(H) & & 0.3894 \textcolor{red}{(-0.0161)} & \textbf{0.3712} \textcolor{ForestGreen}{(+0.0108)} & 0.1696 \textcolor{red}{(-0.0021)} & 0.1019 \textcolor{red}{(+0.0166)} & \textbf{0.7288} \textcolor{ForestGreen}{(+0.0049)} & 0.6405 \textcolor{ForestGreen}{(+0.0291)} & \textbf{0.3849} \textcolor{ForestGreen}{(+0.0085)} & 0.0692\textcolor{red}{(+0.0095)} \\
     
     T5(HD) & & 0.3961 \textcolor{red}{(-0.0094)} & 0.3675 \textcolor{ForestGreen}{(+0.0071)} & 0.1712 \textcolor{red}{(-0.0005)} & \textbf{0.0842} \textcolor{ForestGreen}{(-0.0011)} & 0.7253 \textcolor{ForestGreen}{(+0.0014)} & \textbf{0.6501} \textcolor{ForestGreen}{(+0.0387)} & 0.3759 \textcolor{red}{(-0.0005)} & \textbf{0.0642 }\textcolor{red}{(+0.0045)} \\

\midrule
    
    Qwen & & \textbf{0.4200} & 0.3252 & \textbf{0.1694} & 0.0824 & \textbf{0.7985} & 0.6089 & 0.4357 & 0.0579 \\
 
 Qwen(D) & & 0.3859 \textcolor{red}{(-0.0341)} & 0.3136 \textcolor{red}{(-0.0116)} & 0.1497 \textcolor{red}{(-0.0197)} & \textbf{0.0502 }\textcolor{ForestGreen}{(-0.0322)} & 0.7842 \textcolor{red}{(-0.0143)} & 0.6263 \textcolor{ForestGreen}{(+0.0174)} & 0.4210 \textcolor{red}{(-0.0147)} & \textbf{0.0442} \textcolor{ForestGreen}{(-0.0137)} \\
 
 Qwen(H) & & 0.3924 \textcolor{red}{(-0.0276)} & 0.3397\textcolor{ForestGreen}{(+0.0145)} & 0.1535 \textcolor{red}{(-0.0159)} & 0.0961 \textcolor{red}{(+0.0137)} & 0.7794 \textcolor{red}{(-0.0191)} & 0.6596 \textcolor{ForestGreen}{(+0.0507)} & \textbf{0.4472} \textcolor{ForestGreen}{(+0.0115)} & 0.0492 \textcolor{ForestGreen}{(-0.0087)} \\

Qwen(HD) & & 0.3997 \textcolor{red}{(-0.0203)} & \textbf{0.3457} \textcolor{ForestGreen}{(+0.0205)} & 0.1530 \textcolor{red}{(-0.0164)} & 0.0724 \textcolor{ForestGreen}{(-0.0100)} & 0.7931 \textcolor{red}{(-0.0054)} & \textbf{0.6748} \textcolor{ForestGreen}{(+0.0659)} & 0.4298 \textcolor{red}{(-0.0059)} & 0.0446 \textcolor{ForestGreen}{(-0.0133)} \\

\midrule
    
    Gemma & & 0.3627 & 0.3121 & \textbf{0.1657} & 0.0754 & \textbf{0.7613} & 0.6282 & 0.3882 & 0.0657 \\
 
 Gemma(D) & & 0.3794\textcolor{ForestGreen}{(+0.0167)} & 0.3272 \textcolor{ForestGreen}{(+0.0151)} & 0.1411 \textcolor{red}{(-0.0246)} & \textbf{0.0693} \textcolor{ForestGreen}{(-0.0061)} & 0.7377 \textcolor{red}{(-0.0236)} & 0.6557 \textcolor{ForestGreen}{(+0.0275)} & 0.3831 \textcolor{red}{(-0.0051)} & \textbf{0.0479} \textcolor{ForestGreen}{(-0.0178)} \\
 
 Gemma(H) & & 0.3696 \textcolor{ForestGreen}{(+0.0069)} & 0.3587 \textcolor{ForestGreen}{(+0.0466)} & 0.1603 \textcolor{red}{(-0.0054)} & 0.0909 \textcolor{red}{(+0.0155)} & 0.7557 \textcolor{red}{(-0.0056)} & \textbf{0.6928} \textcolor{ForestGreen}{(+0.0646)} & \textbf{0.3995} \textcolor{ForestGreen}{(+0.0113)} & 0.0631 \textcolor{ForestGreen}{(-0.0026)} \\

Gemma(HD) & & \textbf{0.3863} \textcolor{ForestGreen}{(+0.0236)} & \textbf{0.3601} \textcolor{ForestGreen}{(+0.0480)} & 0.1609 \textcolor{red}{(-0.0048)} & 0.1007 \textcolor{red}{(+0.0253)} & 0.7323 \textcolor{red}{(-0.0290)} & 0.6625\textcolor{ForestGreen}{(+0.0343)} & 0.3918 \textcolor{ForestGreen}{(+0.0036)} & 0.0644 \textcolor{ForestGreen}{(-0.0013)} \\

\midrule
    
    Phi3 & & \textbf{0.3629} & 0.3164 & 0.0937 & 0.0667 & 0.6805 & 0.5629 & 0.3241 & 0.0828 \\
 
 Phi3(D) & & 0.3428 \textcolor{red}{(-0.0201)} & 0.2993 \textcolor{red}{(-0.0171)} & 0.1017 \textcolor{ForestGreen}{(+0.0080)} & \textbf{0.0541} \textcolor{ForestGreen}{(-0.0126)} & 0.6750 \textcolor{red}{(-0.0055)} & 0.5951 \textcolor{ForestGreen}{(+0.0322)} & 0.3309 \textcolor{ForestGreen}{(+0.0068)} & 0.0606 \textcolor{ForestGreen}{(-0.0222)} \\
 
 Phi3(H) & & 0.3601 \textcolor{red}{(-0.0028)} & \textbf{0.3325} \textcolor{ForestGreen}{(+0.0161)} & 0.0976 \textcolor{ForestGreen}{(+0.0039)} & 0.0821 \textcolor{red}{(+0.0154)} & 0.6811 \textcolor{ForestGreen}{(+0.0006)} & \textbf{0.6175} \textcolor{ForestGreen}{(+0.0546)} & 0.3339\textcolor{ForestGreen}{(+0.0098)} & 0.0745 \textcolor{ForestGreen}{(-0.0083)} \\

Phi3(HD) & & 0.3517 \textcolor{red}{(-0.0112)} & 0.3267 \textcolor{ForestGreen}{(+0.0103)} & \textbf{0.1078} \textcolor{ForestGreen}{(+0.0141)} & 0.0681 \textcolor{red}{(+0.0014)} & \textbf{0.6830} \textcolor{ForestGreen}{(+0.0025)} & 0.5976\textcolor{ForestGreen}{(+0.0347)} & \textbf{0.3409} \textcolor{ForestGreen}{(+0.0168)} & \textbf{0.0524} \textcolor{ForestGreen}{(-0.0304)} \\
\bottomrule
\end{tabular}
}
\vspace{-0.3cm}
\end{table*}

\begin{figure}[!t]
    \centering    
    \subfloat[Amazon product review results.]{%
        \includegraphics[width=0.95\columnwidth]{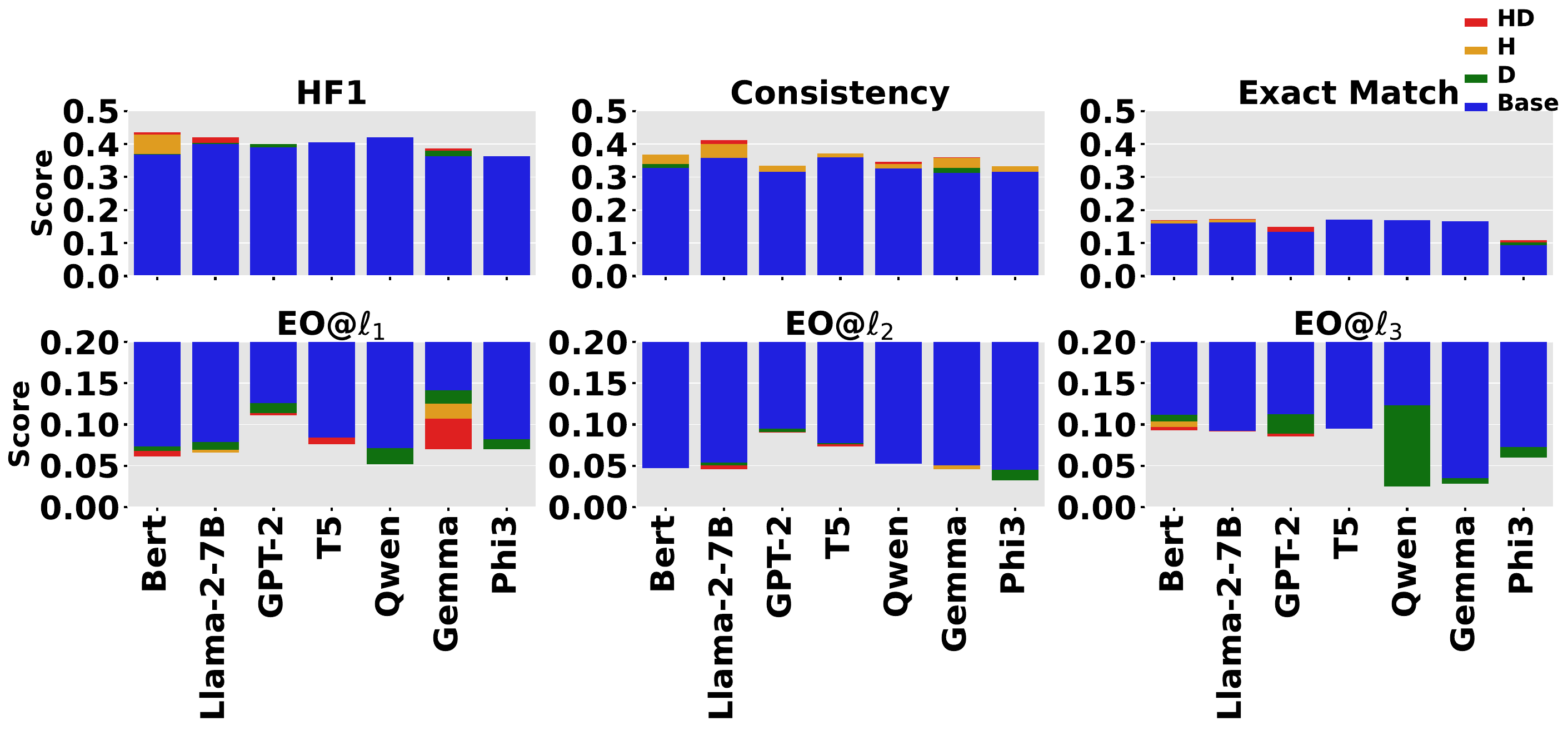}%
        \label{fig:bar_1}%
    }
    \vfill
    \subfloat[DBPedia results.]{%
        \includegraphics[width=0.90\columnwidth]{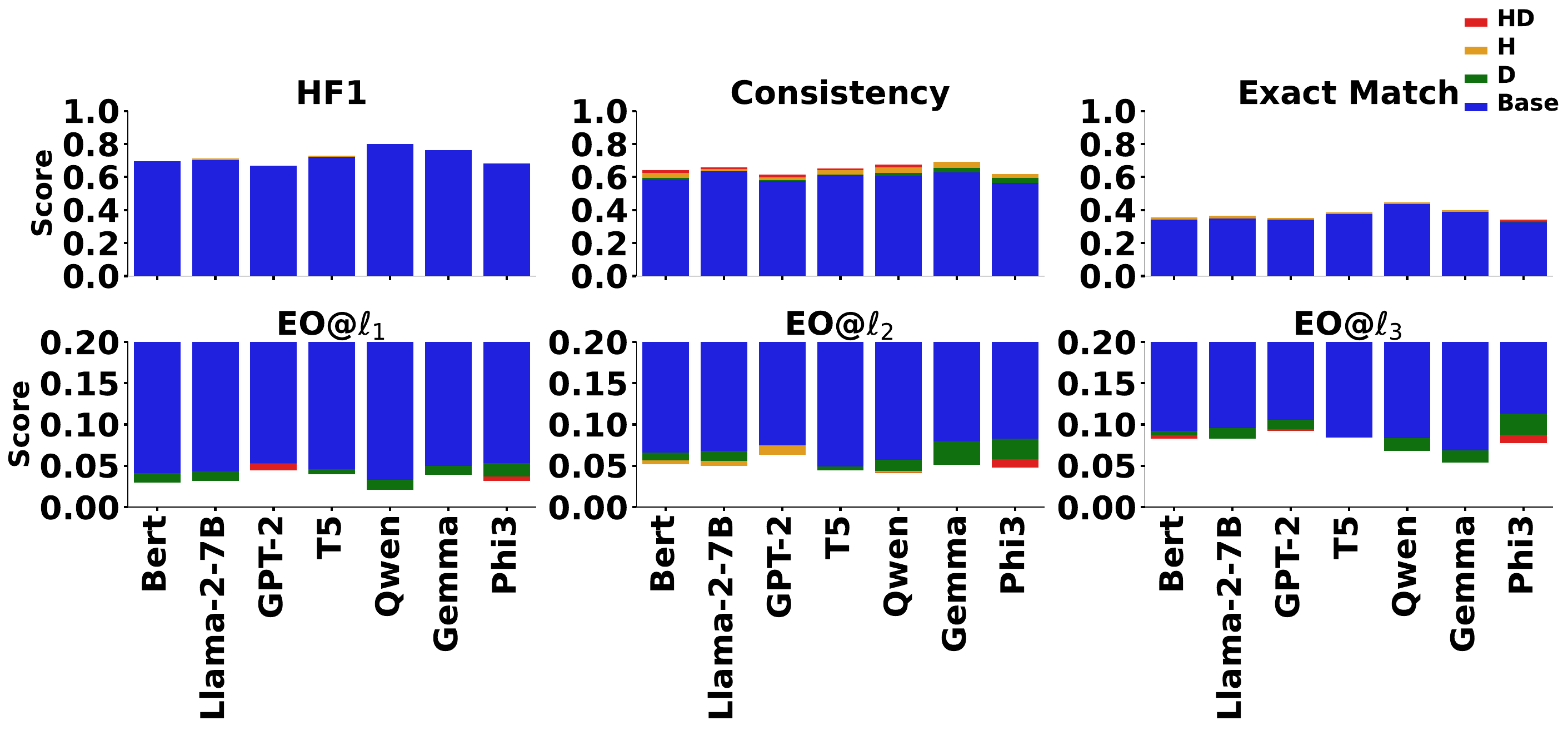}%
        \label{fig:bar_2}%
    }
    \caption{Performance metrics comparison for various models and variants across different evaluation measures. The plots on the top row show metrics where higher values indicate better performance (HF1, Consistency, and Exact Match), whereas the plots on the bottom row (EO@$\ell_1$, EO@$\ell_2$, EO@$\ell_3$) display metrics where lower values are desirable for indicating fairness. The bars for each metric are grouped by model variant, with colors indicating different configurations (Base, D, H, HD). Note the distinct y-axis scales for fairness metrics (EO), highlighting differences in the fairness evaluation across models.}
    \label{fig:facet}
\end{figure}

\noindent\textbf{Evaluation benchmarks and metrics:} For evaluating the MLHC task, we have adopted the Hierarchical F1-Score (HF1- score)~\cite{kosmopoulos2015evaluation}, Consistency, Exact Match and Equalized Odds. Similar to the F1-score, HF1-Score assesses model performance in predicting classes across different hierarchy levels and can be written as 
$$\text{HF1- Score} = \frac{2 \cdot (\text{H-Precision} \cdot \text{H-Recall})}{\text{H-Precision} + \text{H-Recall}}$$. 

H-Recall and H-Precision are analogous to Recall and Precision but evaluate the proportion of correctly predicted classes among all actual/predicted classes. Consistency ensures that predicted labels adhere to hierarchical structures, meaning that predictions across all levels remain within the same hierarchy. Exact Match is a stricter criterion, requiring predictions to not only stay within the hierarchy but also to exactly match true labels at all levels and Equalized Odds assesses fairness by ensuring that the model's predictions are equitable across different demographic groups.


\subsection{Results and Discussion}





We evaluate the performance of the proposed D-TTC layers and analyze the impact of debasing on its effectiveness across various large backbone LLMs. Table \ref{tab:results} presents a detailed comparison of the different LLMs, with and without the inclusion of the model-agnostic TTC layer and Dynamic Reweighting debasing on Amazon product review and DBPedia datasets. The results consistently show that a hierarchical classifier leads to a noticeable performance boost for most backbone LLMs in comparison to the traditional flat classifiers. This demonstrates the effectiveness of the proposed masking layer in addressing hierarchical dependencies. Furthermore, we observe that integrating Dynamic reweighting in a hierarchical classifier not only resulted in better fairness (\textbf{EO}) and consistency but also significantly enhanced the performance of child predictions (shown in appendix \ref{sec:appendix2}), thereby validating the positive impact of encouraging model focus on balanced, sensitive samples and neutral samples, leading to improved predictions in coarse classes. In particular, we can notice that \textbf{HF1-Score} was relatively better across models, which indicates a strong ability to capture hierarchical relationships. We have noticed a slight decline in HF1 for Qwen and T5 for \textit{Amazon Product Review} dataset and all models except T5 and Phi3 for \textit{DBPedia} dataset when TTC and debasing terms are introduced. This suggests that TTC's emphasis on enforcing consistency between layers can result in a trade-off with general performance. However, we have further noticed that the debiasing the hierarchical layer persistently leads to significant improvements in \textbf{Consistency}, \textbf{Exact Match} at each level, highlighting its strength in producing more coherent and fine-grained predictions. These enhancements underline the effectiveness of D-TTC in addressing complex hierarchical classification tasks, ensuring predictions are fair and align better with structured taxonomy.

\begin{figure}[!htb]
    \centering
    \includegraphics[width=0.94\columnwidth]{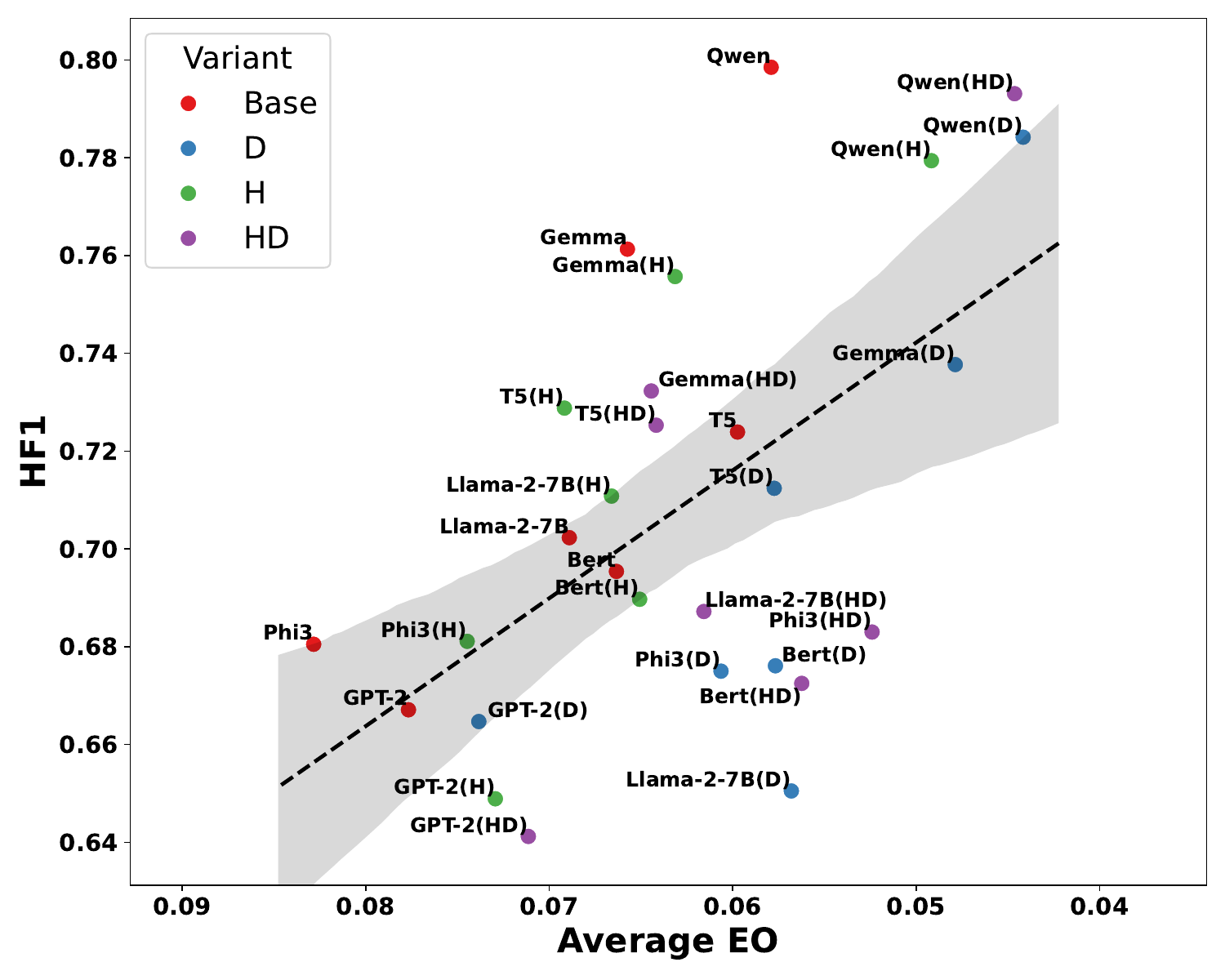}
    \caption{Trade-offs analysis between the HF1 score and Average EO for DBPedia dataset.}
    \label{fig:trade_off_Db}

\end{figure}

Figure \ref{fig:trade_off_Db} illustrates a different correlation trend, primarily because, for DBpedia, the input context is more aligned with the classification tasks. In this case, the model's capability plays a greater role in determining both accuracy and fairness performance, resulting in some models achieving better overall results across both metrics. However, upon closer examination within each group of models, a negative trend remains observable, indicating the underlying trade-offs between performance and fairness.

\begin{figure}[!htb]
    \centering
    \includegraphics[width=0.94\columnwidth]{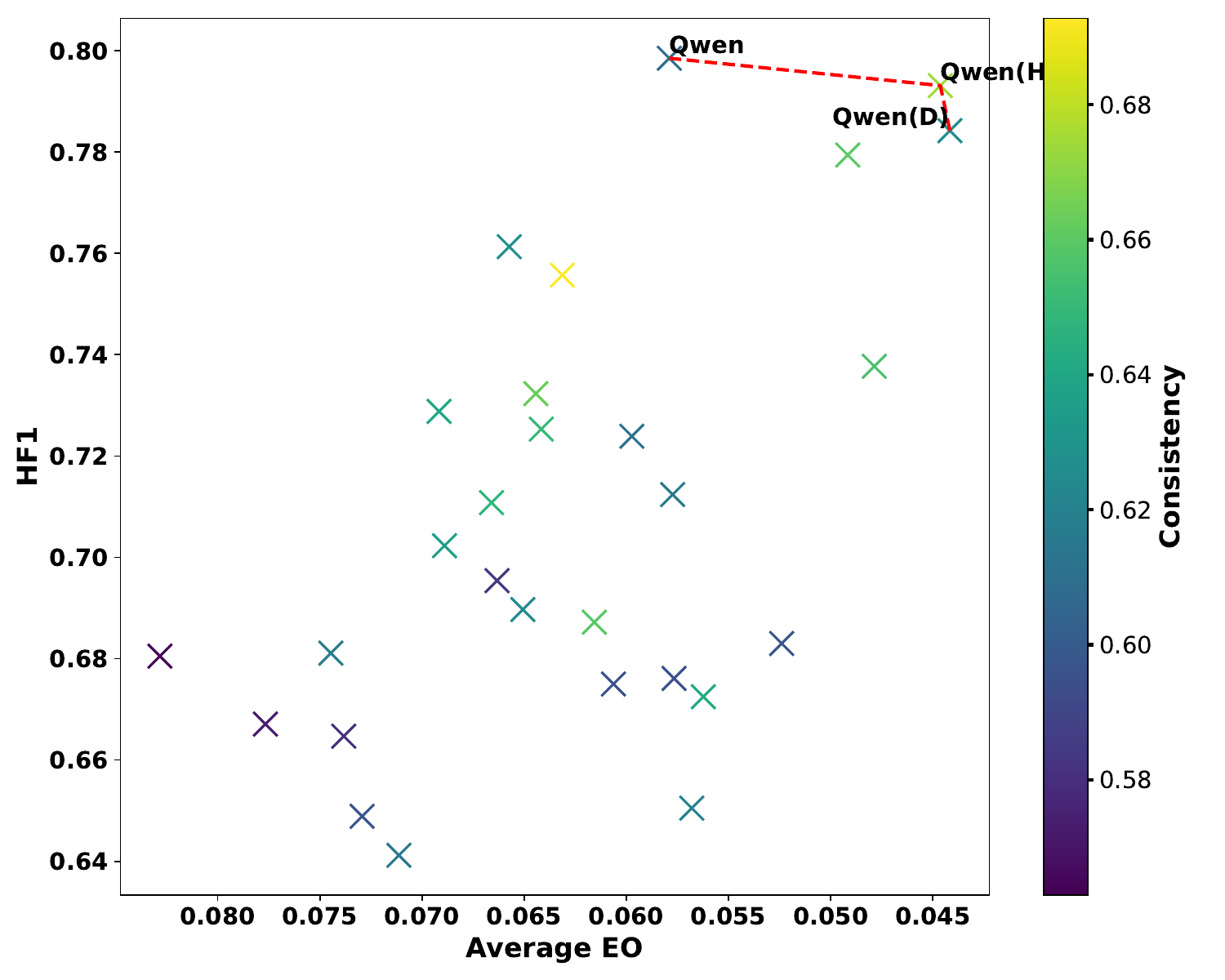}
    \caption{Pareto Front of Equalized Odds (EO) vs. Hierarchical F1 Score (HF1) with Consistency Hue for DBpedia dataset.  The results suggest that Qwen(H) is the best-performing model on the Pareto front.
}
    \label{fig:pareto_Db}

\end{figure}
Figure \ref{fig:pareto_Db} presents all models in terms of their performance across HF1, Consistency, and Average EO. The results suggest that Qwen(H) is the best-performing model on the Pareto front.

\section{Exact match vs Consistency}\label{sec:appendix2}
\begin{figure}[!htb]
    \centering
    \includegraphics[width=0.84\columnwidth]{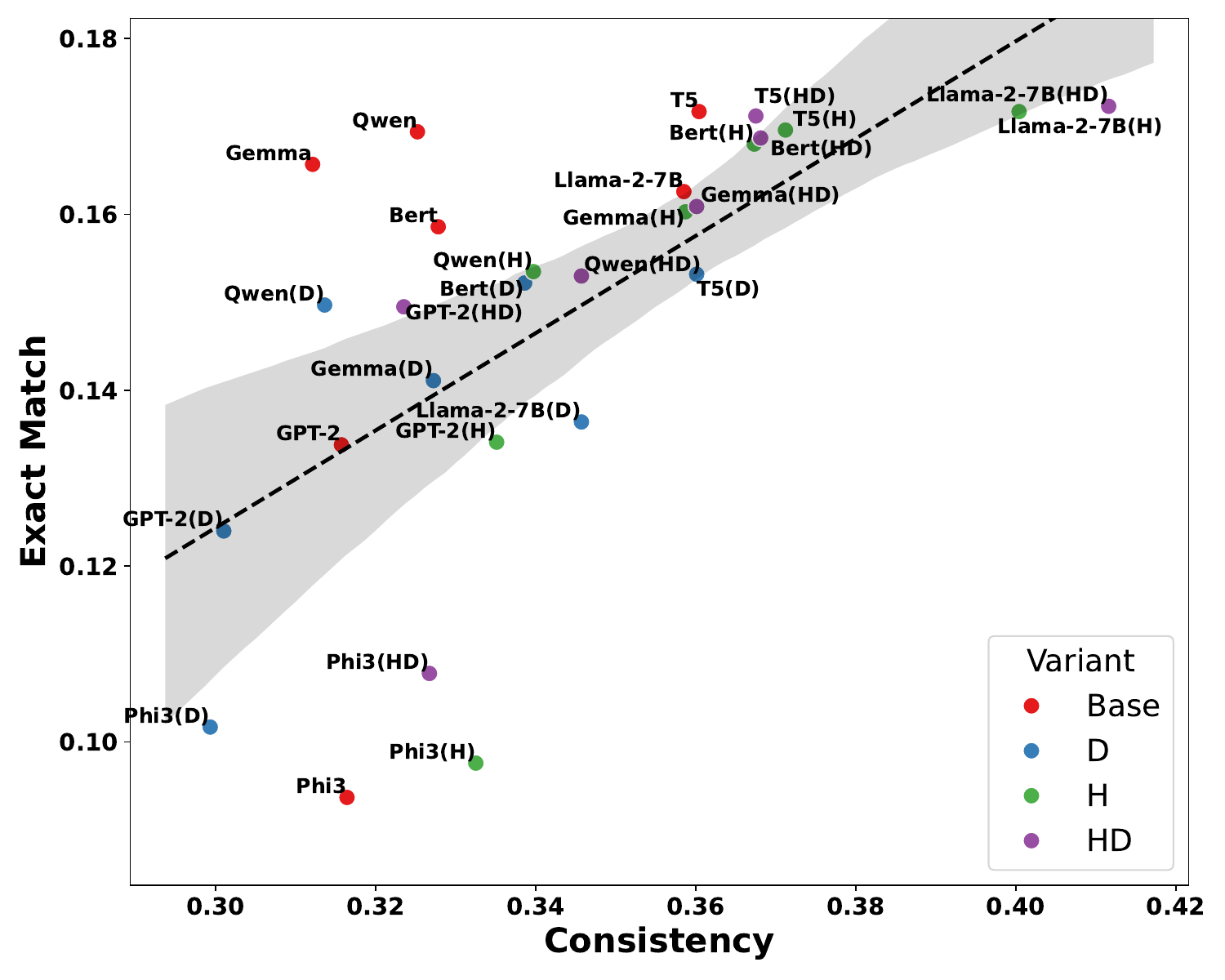}
    \caption{Scatter plot showing the relationship between Exact Match and Consistency for various models on the Amazon Product Review dataset. The models are categorized into different variants (Base, D, H, HD), with a regression line included to highlight the overall trend, showing the strong correlation between the Consistency and Exact match, validating the effectiveness of D-TTC models.}
    \label{fig:con}
\end{figure}

\begin{figure}[!htb]
    \centering
    \includegraphics[width=0.94\columnwidth]{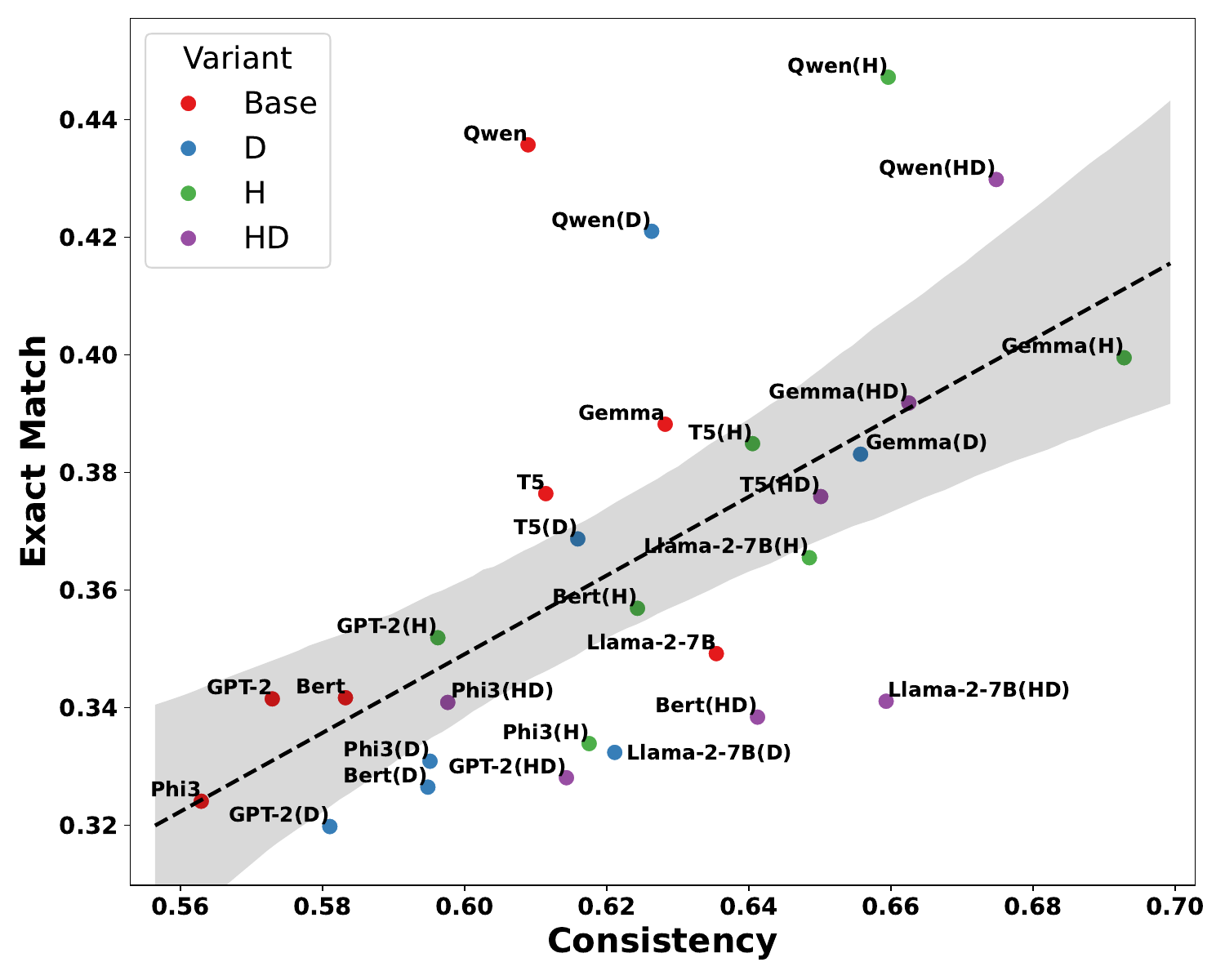}
    \caption{Scatter plot showing the relationship between Exact Match and Consistency for various models on the DBPedia dataset. The models are categorized into different variants (Base, D, H, HD), with a regression line included to highlight the overall trend, showing the strong correlation between the Consistency and Exact match, validating the effectiveness of D-TTC models.}
    \label{fig:con_Db}
\end{figure}

\begin{figure}[!htb]
    \centering
    \includegraphics[height=5.4cm, width=0.95\columnwidth]{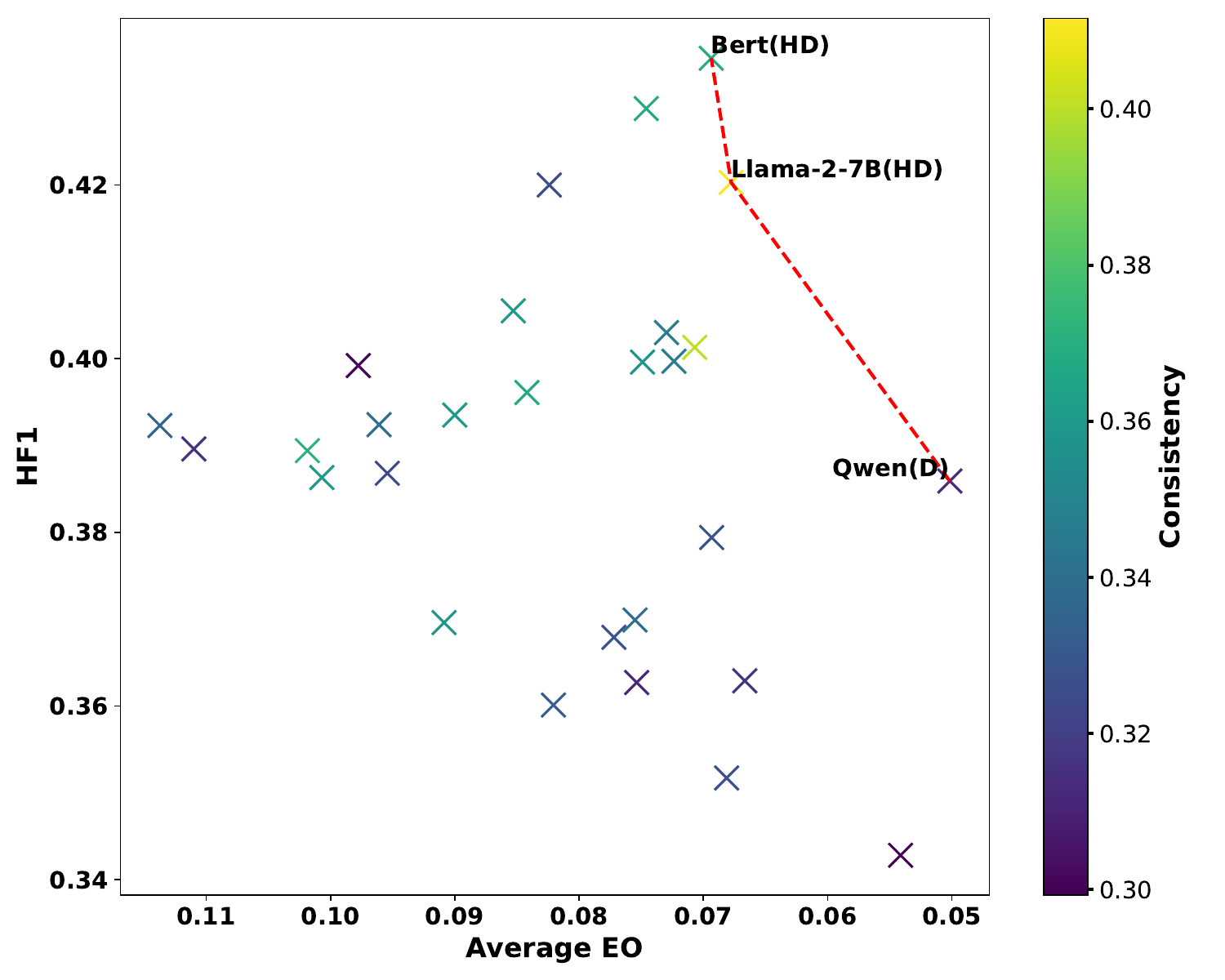}
    \caption{Pareto Front of Equalized Odds (EO) vs. Hierarchical F1 Score (HF1) with Consistency Hue. The scatter plot displays the trade-off between fairness (Average EO) and performance (HF1) across various models. The color of the points represents the consistency of each model, with higher consistency shown in lighter colors. The red dashed line highlights the Pareto front, showcasing the optimal models. Among these, Llama-2-7B(HD) achieves the best balance between fairness and performance, located at the intersection of high consistency and HF1 values.}
    \label{fig:pareto}
   \vspace{-0.49cm}
\end{figure}

\begin{figure}[t]
    \centering
    \includegraphics[height=5.4cm, width=0.96\columnwidth]{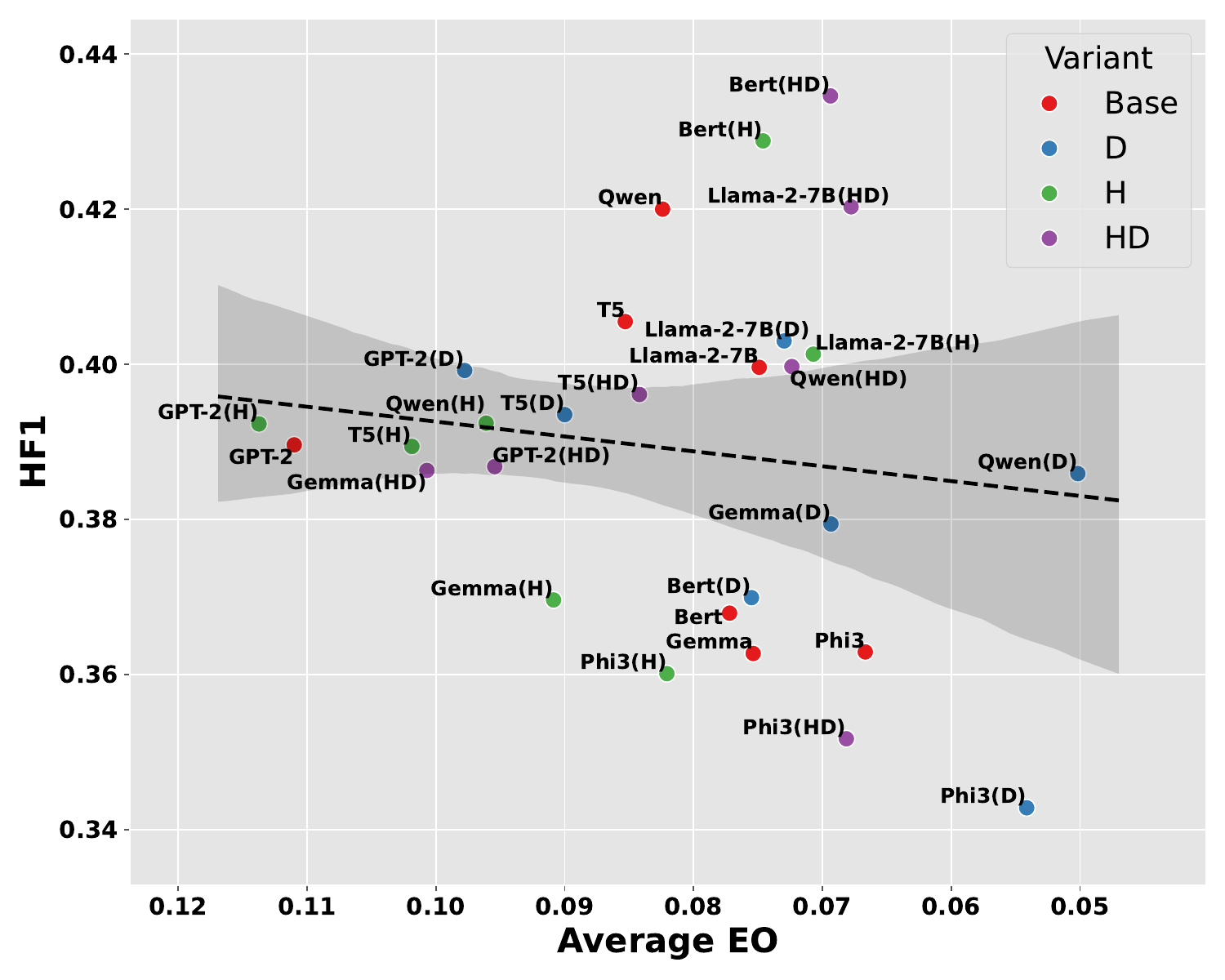}
    \caption{Trade-off between average EO and HF1 score across model variants. This scatter plot visualizes the relationship between the average EO across three levels and the HF1 for various model variants. A regression line (dashed) shows the overall trend with a shaded confidence interval. Model variants are differentiated by color, with labels manually adjusted to avoid overlap. The x-axis is inverted to emphasize lower EO values corresponding to better fairness, highlighting the trade-off between fairness and performance.}
    \label{fig:trade-off}
 \vspace{-0.49cm}
\end{figure}

For further in-depth analysis, we mainly focus on the Amazon dataset. We have studied the trade-offs between the performance metrics (HF1) and fairness metrics(Average EO). As shown in Figure \ref{fig:trade-off}, the scatter plot illustrates the trade-off between \textbf{HF1} and \textbf{Average EO}, where a negative trend is observed both in general and for each group of models. Models that achieve lower EO values (representing better fairness) tend to have reduced HF1 scores, indicating a performance compromise. For example, models like \textit{Phi3(D)} and \textit{Phi3(HD)} prioritize fairness, achieving low EO but with a corresponding drop in HF1. However, models such as \textit{Bert(H)}, \textit{Bert(HD)}, and \textit{Llama-2(HD)} stand out as outliers, achieving both high HF1 scores and relatively low EO values. These models manage to balance performance and fairness more effectively than others, breaking the general trade-off trend. This suggests that these specific model variants may be able to optimize both metrics simultaneously, offering a more favorable trade-off between fairness and performance. Figure \ref{fig:pareto} provides a comprehensive investigation of all the models for the Amazon dataset. The red dashed line illustrates the Pareto front, which highlights the trade-offs between HF1 and Average EO, helping to identify the optimal models. The color of the points corresponds to the consistency level, as indicated by the hue bar, where lighter colors represent higher consistency. From this analysis, we observe that \textit{Llama-2-7B(HD)} stands out as the best combination of high performance (HF1), low bias (Average EO), and relatively high consistency, making it most balanced model along Pareto front.

\section{Conclusion}
In this work, we presented novel D-TTC model agnostic fair masked layer that employs a top-down divide-and-conquer strategy, attending to taxonomy relationships and applying fairness adjustments at each level of the hierarchy by broadcasting fairness and consistency
from parents to children. Experiments conducted on \textit{Amazon Product Review} and \textit{DBPedia} demonstrated significant potential in enhancing the performance of LLMs. Across all evaluated models, the model led to notable improvements in key metrics such as Fairness (EO), Consistency, Exact Match, and HF1 score, as noticed in both Table \ref{tab:results} and Figure \ref{fig:facet}.  Although we observe some trade-offs—such as slight reductions in HF1-Score for certain models (Qwen, T5, and Phi3 on the Amazon Product Review dataset and Gemma, GPT-2, and Qwen on the DBPedia dataset), the overall results reveal substantial gains in fairness, consistency, and exact match. These improvements underscore the efficacy of our D-TTC layer in aligning the model's predictions with the underlying hierarchical structure. The strong positive correlation between Consistency and Exact match (shown in appendix \ref{sec:appendix2}) suggests that our framework can be extended beyond hierarchical tasks to traditional classification problems, where it can serve as a top-down, divide-and-conquer approach to boost performance. 

\section{Limitations}
Overall, the results emphasize the versatility and effectiveness of D-TTC in improving both hierarchical and standard classification tasks across various metrics, particularly in Equalized Odds (EO) and Exact Match, compared to traditional classifiers. This makes it a promising addition to model-agnostic strategies for enhancing LLMs. While TTC-aided LLMs outperform traditional models across multiple metrics and offer broad applicability to classification tasks, however, they depend on a hierarchical data structure and require manual annotation to define class levels. For large-scale datasets with deep hierarchies, this annotation process is labor-intensive, and computing the transition matrix becomes increasingly complex. Additionally, the current approach only accounts for top-down transitions, overlooking bottom-up information that could improve consistency across prediction levels. This limitation hinders the model’s ability to capture relationships between different hierarchy levels. Furthermore, the sequential nature of the TTC framework restricts parallel processing, as predictions must be made in order. This increases computational costs and reduces efficiency, making the method less suitable for real-time applications where speed is critical.

\section{Acknowledgements}

This research is partially supported by the Technology Innovation Institute, Abu Dhabi, UAE. 
\bibliography{custom}  

\end{document}